%% file: acl_latex.tex
\documentclass[11pt]{article}

\usepackage[final]{acl}

\usepackage{booktabs}
\usepackage{tabularx}
\usepackage{makecell}
\usepackage{multirow}
\usepackage{amsmath}
\usepackage{amssymb}
\usepackage{graphicx}
\usepackage[table,xcdraw]{xcolor} 
\usepackage{xspace}
\usepackage{stfloats}
\usepackage[most]{tcolorbox} 
\usepackage{listings}
\usepackage{enumitem} 

\definecolor{boxback}{RGB}{248, 248, 250}
\definecolor{boxframe}{RGB}{100, 100, 100}

\newcommand{\code}[1]{\texttt{#1}}

\newtcolorbox{columnprompt}[1][]{
    breakable,
    enhanced,
    colback=boxback,
    colframe=boxframe,
    coltitle=white,
    fonttitle=\bfseries\scriptsize\sffamily,
    title={Lifter Agent System Prompt},
    boxrule=0.5pt,
    arc=1mm,
    left=4pt, right=4pt, top=4pt, bottom=4pt,
    fontupper=\scriptsize\raggedright,
    #1
}

\setlist[itemize]{leftmargin=*, nosep, noitemsep, topsep=2pt, parsep=2pt}
\setlist[description]{leftmargin=1em, nosep, noitemsep, topsep=2pt, parsep=2pt, font=\bfseries}

\usepackage{comment}
\usepackage{times}
\usepackage{latexsym}
\newenvironment{revision}{}{}
\newcommand{\fixme}[1]{#1}

\newcommand{\proj}[1]{\textit{CuBridge}}
\newcommand{\irname}{CuIR\xspace}
\newcommand{\gc}{\cellcolor{gray!10}}

\usepackage[T1]{fontenc}

\usepackage[utf8]{inputenc}

\usepackage{microtype}

\usepackage{inconsolata}

\usepackage{graphicx}

\graphicspath{{figures/}}

\title{CuBridge: An LLM-Based Framework for Understanding and Reconstructing High-Performance Attention Kernels}

\author{
  \textbf{Xing Ma}$^{1,2\ast}$, \textbf{Yangjie Zhou}$^{3\ast\dagger}$, \textbf{Wu Sun}$^1$, \textbf{Zihan Liu}$^{1,2}$, \textbf{Jingwen Leng}$^{1,2}$, \\
  \textbf{Yun Lin}$^{1\dagger}$, \textbf{Shixuan Sun}$^1$, \textbf{Minyi Guo}$^1$, \textbf{Jin Song Dong}$^3$ \\
  $^1$Shanghai Jiao Tong University, $^2$Shanghai Qi Zhi Institute, $^3$National University of Singapore \\
  \texttt{\{maxing1207, by5892q, altair.liu, lin\_yun, sunshixuan\}@sjtu.edu.cn} \\
  \texttt{\{leng-jw, guo-my\}@cs.sjtu.edu.cn}, \texttt{\{yj\_zhou, dcsdjs\}@nus.edu.sg}
}

\begin{document}
\maketitle

\def\thefootnote{$\ast$}\footnotetext{Equal contribution.}\def\thefootnote{\arabic{footnote}}
\def\thefootnote{$\dagger$}\footnotetext{Corresponding authors.}\def\thefootnote{\arabic{footnote}}

\input{latex/abstract}
\input{latex/introduction}

\input{latex/preliminaries}

\input{latex/methods}

\input{latex/experiment}

\input{latex/conclusion}

\input{latex/limitations}
\input{latex/acknowledgements}

\bibliography{latex/refs}

\clearpage
\appendix
\input{latex/appendix}


\end{document}

%% file: latex/abstract.tex
\begin{abstract}

Efficient CUDA implementations of attention mechanisms are critical to modern deep learning systems, yet supporting diverse and evolving attention variants remains challenging.
Existing frameworks and compilers trade performance for flexibility, while expert-written kernels achieve high efficiency but are difficult to adapt.
Recent work explores large language models (LLMs) for GPU kernel generation, but prior studies report unstable correctness and significant performance gaps for complex operators such as attention.

We present \proj{}, an LLM-based framework that adapts expert-written attention kernels through a structured lift–transfer–lower workflow. \proj{} starts from expert-written CUDA attention kernels and lifts them into an executable intermediate representation that makes execution orchestration explicit while abstracting low-level CUDA syntax. Given a user-provided PyTorch specification, \proj{} generates and verifies a target IR program, then reconstructs optimized CUDA code via reference-guided lowering.
Across diverse attention variants and GPU platforms, \proj{} consistently produces correct kernels and substantially outperforms general frameworks, compiler-based approaches, and prior LLM-based methods.

\end{abstract}

%% file: latex/introduction.tex
\section{Introduction}

Attention kernels are a central performance component in modern deep learning systems~\cite{pytorch,TensorFlow,TVM}.
As model architectures evolve, beyond standard softmax attention, attention increasingly appears in customized forms, including new masking behaviors, modified score computations, alternative normalization rules, and fused pipelines.
Efficiently supporting these variants  on GPUs is therefore not a one-off engineering task, but an ongoing systems challenge.

Existing approaches expose a fundamental trade-off between generality and performance.
High-level frameworks~\cite{pytorch} and compilers~\cite{FlexAttention} provide flexible expression of new attention semantics but rely on generic and inefficient implementations, while expert libraries~\cite{FlashAttention-2,FlashAttention-3} achieve strong efficiency through carefully engineered kernels that are difficult to adapt.
As a result, extending high-performance attention kernels to new variants often requires substantial manual effort.

Large language models (LLMs) offer a promising direction for reducing the manual effort involved in GPU kernel development, and recent benchmarks have explored their use for kernel generation~\cite{Kernelbench,tritonbench}.
However, for complex operators such as attention, prior studies report unstable correctness and substantial performance gaps relative to optimized baselines~\cite{Kernelbench}, indicating that directly applying LLMs to attention kernels remains unreliable in practice.

A key observation underlying this work is that expert-written CUDA kernels already encode correct and efficient execution behavior.
Rather than generating attention kernels from scratch, we treat these expert implementations as performance references and focus on enabling controlled semantic adaptation while preserving their execution structure.
This perspective motivates a workflow that separates semantic reasoning from low-level code manipulation, allowing modifications to be expressed at an appropriate level of abstraction.

Guided by this insight, we present \proj{}, an LLM-based framework that adapts expert-written attention kernels through a structured lift-transfer-lower workflow.
\proj{} first lifts source CUDA code into a structured intermediate representation \irname, which makes execution orchestration explicit while retaining performance-critical information.
It then transfers the lifted IR program to match user-specified attention semantics, and finally lowers the transformed representation back into optimized CUDA code via reference-guided reconstruction.
Crucially, the intermediate representation is executable, enabling verification at each stage and ensuring semantic correctness.

We implement and evaluate \proj{} on NVIDIA A100 and H100 across a range of attention variants and real-model configurations. 
Our results show that \proj{} consistently produces correct kernels and achieves, on average, speedups of $16.03\times$ over general frameworks, $1.39\times$ over template-based compilers, and $3.33\times$ over prior LLM-based approaches.
These results demonstrate the effectiveness of our approach for supporting evolving attention mechanisms with both correctness and high performance.

Our contributions are summarized as follows:
\begin{itemize}[leftmargin=*, nosep]
\item We propose a new LLM-assisted paradigm for adapting expert-written attention kernels that avoids end-to-end code generation, and instead performs semantic adaptation through a structured lift--transfer--lower workflow.

\item We design \irname{}, an executable intermediate representation that makes performance-critical execution orchestration explicit while abstracting low-level CUDA syntax, and build an end-to-end system \proj{} on top of it to enable reliable semantic adaptation, intermediate verification, and reference-guided kernel reconstruction.
\item We evaluate \proj{} across a wide range of attention variants and GPU platforms, showing that it consistently produces correct kernels and outperforms general frameworks, template-based compilers, and prior LLM-based approaches.

\end{itemize}

%% file: latex/preliminaries.tex
\section{Preliminaries}

The goal of this work is to efficiently support high-performance CUDA~\cite{nvcc} kernels for diverse attention variants.
Achieving high performance on GPUs relies on complex and carefully engineered coordination across multiple levels of parallelism and memory hierarchies~\cite{a100_whitepaper,h100_whitepaper}.
As attention mechanisms continue to evolve, supporting new semantics with high-performance CUDA kernels becomes increasingly challenging.

\subsection{GPU Architecture and Execution Model}

\begin{figure}[t] 
    \centering
    \includegraphics[width=0.8\columnwidth]{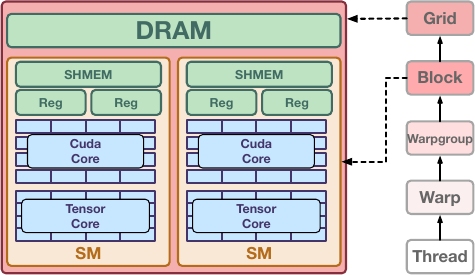} 
    \caption{Hardware architecture and execution model of modern GPUs.}
    \label{fig:gpu} 
    \vspace{-0.5cm}
\end{figure}

Modern GPUs expose a hierarchical organization of both execution units and memory resources, as shown in Figure~\ref{fig:gpu}.
CUDA kernels run under the SIMT model~\cite{SIMT}, where threads are grouped into warps and scheduled in thread blocks, while data resides in a memory hierarchy including global memory, shared memory, and registers.
High-performance CUDA kernels depend on carefully coordinating computation and data movement across these execution units and memory levels.
We refer to this coordination as \emph{execution orchestration}, which determines how work is assigned to parallel units, how operations are ordered with synchronization, and how data is moved across the memory hierarchy, all of which directly impact performance.

Modern hardware features such as tensor cores~\cite{v100_whitepaper}, asynchronous memory copies~\cite{a100_whitepaper}, and pipelined execution ~\cite{h100_whitepaper} further increase the complexity of execution orchestration.
For example, warp-specialized~\cite{cutlass,warp_specialization} pipelines assign different roles to different warps within a thread block and rely on explicit synchronization to coordinate data movement and computation.
As a result, execution orchestration plays a central role in determining the performance characteristics of CUDA kernels.

\subsection{Attention Variants}

Given query, key, and value matrices $Q, K, V \in \mathbb{R}^{L \times d}$, the standard scaled dot-product attention~\cite{Attentionqkv} is defined as:
\begin{equation}
    O = \operatorname{Softmax}\left( \frac{QK^T}{\sqrt{d}} \right)V
\end{equation}

A wide range of attention variants have been proposed to support different modeling requirements.
At a semantic level, many of these variants can be understood as modifying different semantic components of the attention computation.
In particular, attention can be viewed as a sequence of score computation, masking, and normalization:
\begin{equation}
    O = \operatorname{\textbf{Norm}}\left( \operatorname{\textbf{Mask}}\left( \operatorname{\textbf{Score}}(\frac{QK^T}{\sqrt{d}}) \right) \right) V
\end{equation}
Under this view, variants commonly modify the score function ~\cite{Relative_Position}, the masking rules ~\cite{FlashMask,Exploring_the_Limits,Longformer,Leveraging_Passage_Retrieval,Ring_Attention}, or the normalization behavior ~\cite{FlashAttention,Sigmoid_Self-Attention}.

While these semantic changes appear localized at the algorithmic level, their realization in GPU kernels often requires conditional execution, boundary handling, or changes in loop structure.
Such control-flow and data-dependence changes directly affect how computation, synchronization, and data movement are orchestrated on the GPU, making correct and efficient kernel implementation significantly more challenging.

\subsection{Current Support for Attention Variants}

\begin{figure}[t] 
    \centering
    \includegraphics[width=\columnwidth]{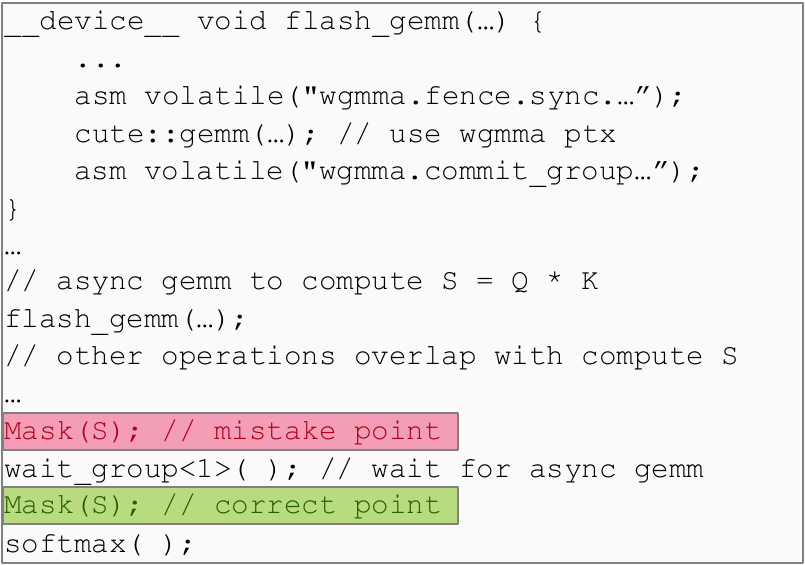} 
    \caption {Failure case of directly applying LLM to modify FlashAttention-3 CUDA code for the PrefixLM mask variant with GPT-5.}
    \label{fig:mistake} 
\end{figure}

Existing systems for supporting attention variants exhibit a fundamental trade-off between generality and performance.
General frameworks (e.g., PyTorch~\cite{pytorch}) allow new attention semantics to be expressed easily at a high level, but do not provide expert-level CUDA kernels for most variants.
Template-based compilation approaches (e.g., FlexAttention~\cite{FlexAttention}) allow limited customization within fixed templates, while expert libraries (e.g., FlashAttention~\cite{FlashAttention-2, FlashAttention-3}) achieve high efficiency but require substantial manual effort to extend.
As a result, non-standard attention variants either suffer from noticeable performance degradation or demand significant manual engineering.

Recent work has explored the use of large language models (LLMs) to automate GPU kernel generation.
Benchmark studies~\cite{Kernelbench} show that while LLM-generated CUDA kernels can perform competitively on simple operators, their correctness becomes unstable and performance degrades significantly for complex kernels such as attention, with slowdowns of up to 34.9$\times$ relative to optimized baselines.
Several recent efforts explore LLM-based code generation and optimization, including multi-agent collaboration ~\cite{CUDA-LLM,MULTIAGENT-AGENTCODER,MULTIAGENT-PAIRCODER} , profiler-driven feedback ~\cite{DBLP:STARK,MULTIAGENT-MAPCODER, Self-Debug}, and natural language-based planning ~\cite{QiMeng-Attention}.
Despite their promise, these approaches do not yet provide a reliable way to realize expert-level execution orchestration for complex kernels.

A natural attempt to improve flexibility is to directly modify expert-written CUDA kernels using LLMs.
However, as shown in Figure~\ref{fig:mistake}, expert kernels often rely on complex syntax and implicit asynchronous execution.
Even when the intended semantic modification is clear, locating the correct code region to update can be error-prone, making direct code-level adaptation unreliable.

%% file: latex/methods.tex
\section{Method}
\label{sec:method}

\begin{figure}[t]
    \centering
    \includegraphics[width=\columnwidth]{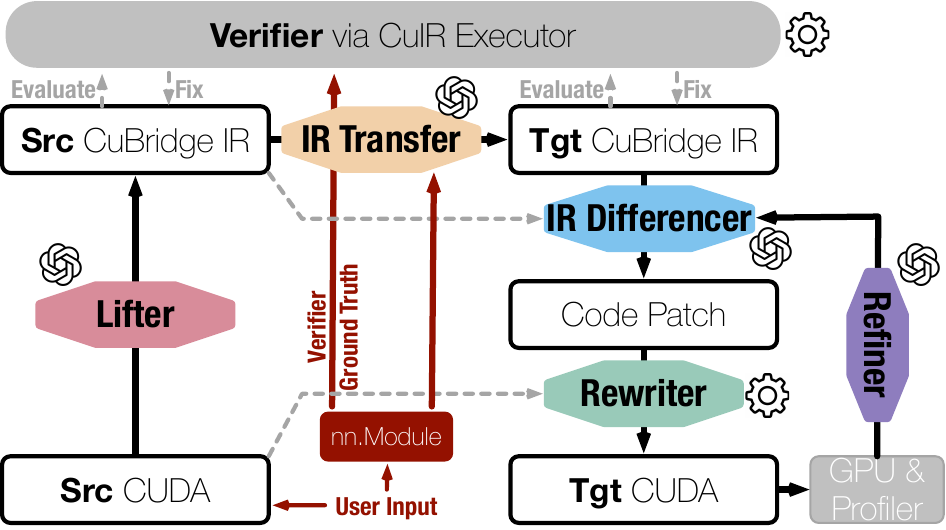} 
    \caption{Overview of \proj{}.}
    \label{fig:overview} 
    \vspace{-1em} 
\end{figure}

\subsection{Overview}
We present \proj{}, an LLM-powered framework for adapting expert-written CUDA attention kernels to new algorithmic variants while preserving high-performance execution orchestration.
Instead of generating kernels from scratch,
\proj{} follows a lift--transfer--lower workflow centered on an executable intermediate representation, \irname{}, which supports both understanding and reconstruction of expert CUDA kernels.

As shown in Figure~\ref{fig:overview}, given a high-performance source CUDA kernel and a user-provided PyTorch reference specifying target semantics, \proj{} proceeds in three stages.
(1) \textbf{Lifting:} \proj{} lifts the Source kernel into \irname{}, which abstracts away low-level syntactic details while retaining performance-critical execution orchestration.
(2) \textbf{Transform:} guided by Target semantics, \proj{} generates  the Target \irname from the lifted Source \irname{}, jointly reasoning about semantic correctness and performance optimization.
(3) \textbf{Lowering:} \proj{} lowers the Target \irname{} back to CUDA by differencing the Source and Target \irname{} programs and generating minimal patches that update the corresponding code regions in the Source kernel to reconstruct the target kernel.

\begin{figure*}[th] 
    \centering
    \includegraphics[width=0.95\textwidth]{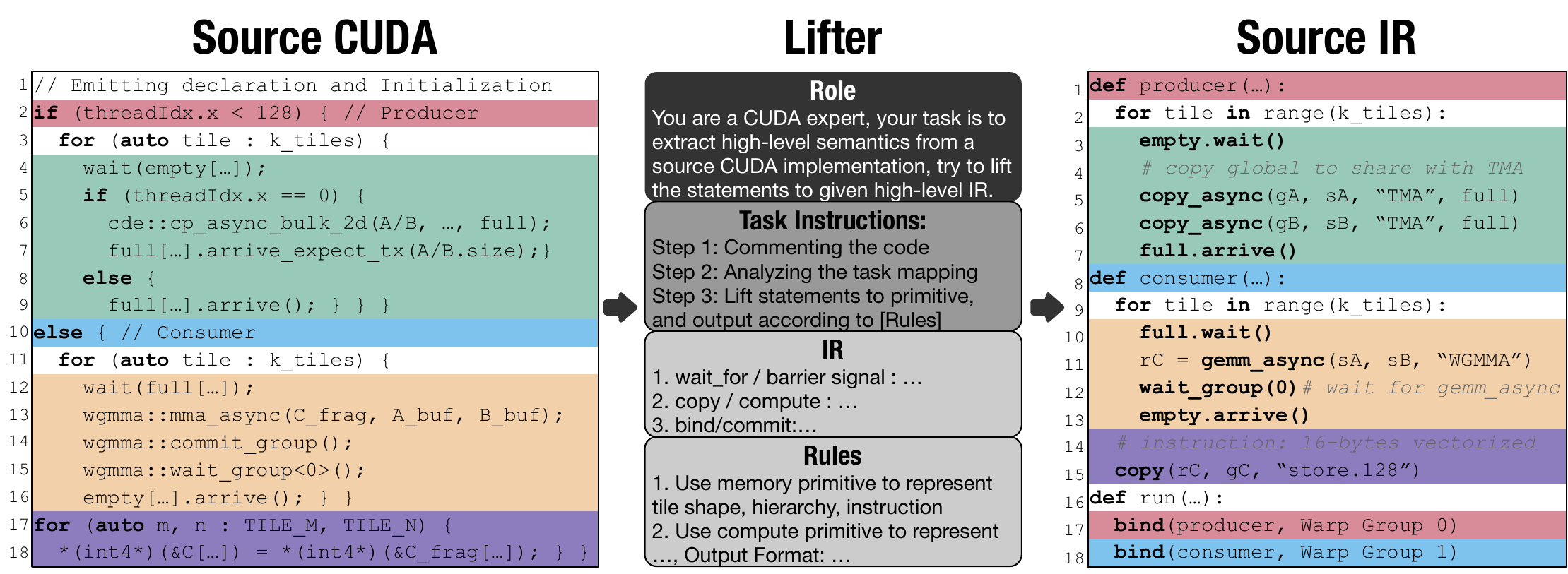}
    \vspace{-0.1cm}
    
    \caption{\textbf{The Semantic Lifting Process.} The left panel shows the complex Source CUDA, and the right panel displays the lifted Source IR, where code blocks of the same color represent mapping relationships. The center shows the modularized prompt, consisting of the CuIR documentation and the Chain-of-Thought reasoning process.}
    \vspace{-0.3cm}
    
    \label{fig:code_detail} 
\end{figure*}

\subsection{\irname Design}
\label{sec:ir_definition}

\begin{table}[t]
    \centering
    \scriptsize 
    \renewcommand{\arraystretch}{1.2} 
    \setlength{\tabcolsep}{4pt}
    
    \newcolumntype{L}{>{\raggedright\arraybackslash}X}
    
    \begin{tabularx}{\columnwidth}{l L L}
        \toprule
        \textbf{Category} & \textbf{Primitives} & \textbf{Exposed Information} \\
        \midrule
        \textbf{Memory} & \texttt{alloc}, \texttt{copy}, \texttt{copy\_async} & Tile Shape, Memory Hierarchy \\
        \cmidrule(lr){1-3} 
        \textbf{Compute} & \texttt{gemm\_async}, \texttt{gemm} & Instruction Selection, Operand \\  
        \cmidrule(lr){1-3}
        \textbf{Sync.} & \texttt{barrier.\{wait, arrive\}} & Data Dependency \\
        \cmidrule(lr){1-3}
        \textbf{Control} & \texttt{bind}, \texttt{commit} & Execution Granularity \\
        \bottomrule
    \end{tabularx}

    \caption{\irname Primitives and Exposed Hardware Semantics.}
    \label{tab:ir_primitives}
    
    \vspace{-1em} 
\end{table}

To enable effective LLM-based understanding and reconstruction of expert CUDA kernels, we design \irname{}, a Pythonic and executable intermediate representation that explicitly captures execution orchestration.
Structurally, \irname{} is expressed in Python syntax and built on a set of custom primitives that operate on data tiles as PyTorch tensors.

\noindent\textbf{Design Rationale.} 
High-performance CUDA kernels derive their efficiency from carefully structured execution orchestration, rather than from individual instructions in isolation.
Accordingly, \irname{} is designed to explicitly preserve three classes of information: the operations invoked by the kernel, the ordering and data dependencies among these operations, and the execution granularity and synchronization scope at which they are performed.
As summarized in Table~\ref{tab:ir_primitives}, Memory and Compute primitives make operations explicit, including tile-level data movement and instruction-level choices, while Synchronization and Control primitives encode dependencies, execution ordering, and collective execution granularity.
These preserved elements are directly linked to correctness and performance, as GPU efficiency hinges on latency hiding and data locality.
Conversely, \irname{} abstracts away low-level implementation details such as fragmented thread-level indexing and verbose CUDA syntax, which are not essential to reasoning about execution structure.

\noindent\textbf{Key Capabilities.} 
\irname{} provides two capabilities that are essential to \proj{}.

\noindent\textit{1) Kernel Understanding and Reconstructing.} 
By making execution orchestration explicit through structured primitives, \irname{} exposes the execution logic of a CUDA kernel as a composition of well-defined operations and dependencies, rather than low-level syntax. This structured representation allows an LLM to understand and reason kernel behavior, enabling semantic changes to be captured in the generated Target \irname{} and localized via IR differencing for Target CUDA reconstruction.

\noindent\textit{2) Intermediate Verification.} 
\irname{} is designed to be executable and verifiable through a dedicated IR backend executor, enabling validation of algorithmic logic and data dependencies at the IR level.
To ensure semantic alignment with CUDA execution, the executor explicitly models the kernel’s parallel structure and synchronization behavior.
Independent parallel tasks are executed sequentially without affecting correctness, while dependent tasks are ordered to respect synchronization and data-dependence constraints.
This execution model allows IR-level verification to faithfully reflect CUDA kernel behavior.

\subsection{Semantic Lifting}
\label{sec:stage1_lifting}

The first stage of \proj{} lifts an expert-written CUDA kernel into \irname{}, extracting its high-level execution orchestration while abstracting away low-level implementation complexity.
As illustrated in Figure~\ref{fig:code_detail}, the \textit{Lifter} performs a single LLM invocation with structured chain-of-thought (CoT) reasoning, organized into three sub-tasks.

\noindent\textbf{1) Syntax Annotation.} 
This step resolves low-level syntactic ambiguity by annotating CUDA operations with explicit semantic comments.
Expert CUDA kernels often rely on low-level PTX intrinsics or library-specific APIs (e.g., CuTe), whose semantics are implicit in the source code and not directly exposed.
The \textit{Lifter} augments such operations with concise semantic descriptions grounded in official CUDA and CuTe documentation, producing an annotated CUDA representation that makes the intent of low-level instructions explicit.

\noindent\textbf{2) Code-to-Worker Mapping Analysis.}
This step identifies which code regions in the source CUDA kernel are executed by which parallel execution units, making implicit execution relationships explicit.
Here, we use the term \emph{worker} to denote a cooperative CUDA execution unit, such as a thread block, warp group, or warp.
Specifically, the \textit{Lifter} analyzes control-flow predicates over thread indices and attributes guarded code regions to the corresponding workers.
For example, as shown in Figure~\ref{fig:code_detail} (Source CUDA, Line 2), the branch \texttt{if (threadIdx.x < 128)} partitions the kernel into two worker-aligned regions, which are later bound to different warp groups in the generated IR.

\noindent\textbf{3) Primitive Lifting.} 
In the final step, the \textit{Lifter} translates each worker-aligned code region into a sequence of \irname{} primitives and constructs the corresponding Source IR.
This process recovers primitive parameters such as tile shapes, memory placement, instruction variants, and synchronization scope by analyzing index expressions, memory access patterns, and surrounding control structure.
The generated \irname{} program explicitly encodes the execution orchestration of the original kernel and serves as the input to subsequent transformation and lowering stages.

\noindent\textit{Post-Lifting Verification \& Refinement.} 
The generated \irname
program is executed by the backend executor and its output is compared against the source CUDA within a numerical tolerance (e.g., $10^{-2}$ for fp16), following CUTLASS~\cite{cutlass}.
On mismatch, the executor log is used to diagnose errors such as incorrect tile parameters or missing synchronization, and guides the regeneration of the IR.
This closed-loop refinement ensures that the final Source IR faithfully preserves the source kernel semantics.

\subsection{Semantic Transformation}
\label{sec:stage2_transform}

In this stage, the \textit{IR Transfer} transforms the lifted Source IR into a Target IR that implements the target operator semantics specified by a user-provided PyTorch reference, while enabling the performance-critical execution orchestration.

\noindent\textbf{1) Semantic Alignment.} 
The \textit{IR Transfer} first analyzes the PyTorch reference to identify the semantic difference between source IR and target operator. 
Guided by this gap, it maps the required semantics to \irname{} primitives and determines where the source orchestration must be extended or modified.
In this way, the generated Target IR program matches the target computation while remaining structurally aligned with the Source IR.

\begin{figure}[t]
    \centering
    \includegraphics[width=\columnwidth]{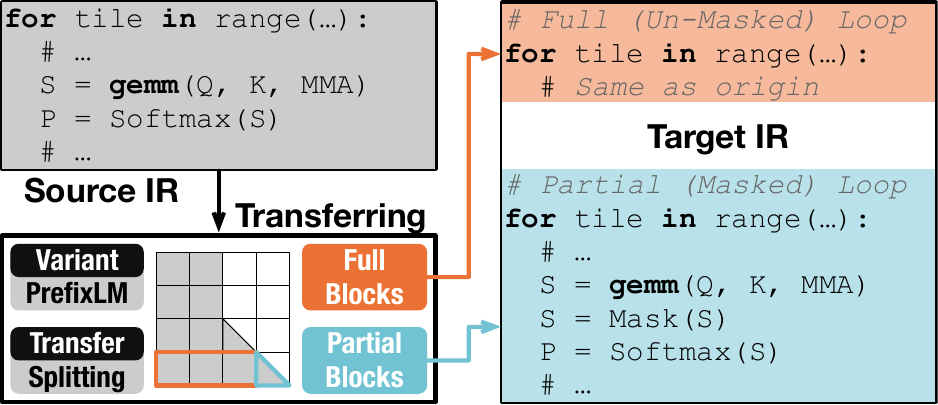} 
    
    \vspace{-0.1cm}
    
    \caption{Performance-Aware Transformation for PrefixLM. The IR Transfer performing Loop Splitting to confine mask checks exclusively to boundary tiles.}
    \vspace{-0.3cm}
    
    \label{fig:ir_transform} 
\end{figure}

\noindent\textbf{2) Performance-Aware Transformation.} 
A semantics-correct Target IR can still be inefficient.
To maintain efficiency, the \textit{IR Transfer} further analyzes the performance implications of the required semantic changes during Target IR generation.
For PrefixLM~\cite{Exploring_the_Limits}, the validity of the attention mask correlates with the nested loop iteration space: element-wise checks are required only for partial blocks, whereas other blocks are fully valid.
As shown in Figure~\ref{fig:ir_transform}, the \textit{IR Transfer} applies loop splitting to construct a check-free \textit{Full Loop} for full blocks and a \textit{Partial Loop} that applies masking for partial blocks, thereby avoiding unnecessary element-wise checking in the full-loop region.

\noindent\textit{Post-Transformation Verification.} 
The generated Target IR is executed by the \irname{} backend executor and validated against the user-provided PyTorch reference implementation, using the same verification procedure as in Sec.~\ref{sec:stage1_lifting}. Any mismatch triggers iterative refinement.

\subsection{Kernel Reconstruction}
\label{sec:stage3_reconstruction}

The final stage lowers the transformed Target IR into high-performance Target CUDA code.
\proj{} reconstructs the target implementation by applying localized, IR-guided code patches within a structured ReAct workflow~\cite{ReAct}.

\noindent\textbf{1) Differential Analysis.} 
Reconstruction begins by comparing the Source IR and the Target IR to identify semantic differences introduced during transformation.
Leveraging the region-level correspondence established in the lifting stage, these IR-level changes are traced back to their corresponding locations in the Source CUDA code, precisely identifying the code regions that require modification.

\noindent\textbf{2) Reference-Guided Lowering.} 
To translate the Target \irname{} program into CUDA code, \proj{} performs reference-guided lowering by using the \textit{Source CUDA} kernel as an implementation reference for the Source IR.
This reference specifies how abstract IR primitives should be concretely realized in CUDA, ensuring consistency with the original kernel’s implementation style.
The lowering process focuses on two tasks:
\begin{itemize}[leftmargin=*, nosep]
    \item \textbf{Primitive Lowering:} Lowering abstract IR primitives (e.g., \texttt{copy\_async}) to concrete hardware intrinsics (e.g., \texttt{cp.async.ca}) used in the source kernel.
    \item \textbf{Index Mapping:} Expanding tile-level operations into thread-level indexing that follows the data layout and indexing patterns in the source kernel.
\end{itemize}

\noindent\textbf{3) Iterative Patching.}
Finally, \textit{the Rewriter} reconstructs the target CUDA kernel by applying localized and minimal code patches to the source implementation.
These updates are realized through a line-level editing action (\texttt{Edit\_Line}) that supports insertion, deletion, and modification of code lines within identified regions.
By restricting changes to affected code segments, this patch-based reconstruction ensures faithful alignment between the Target IR and CUDA code, while avoiding the context-length limitations of full-file rewriting.

%% file: latex/experiment.tex
\begin{figure*}[th]
    \centering
    \includegraphics[width=0.99\textwidth]{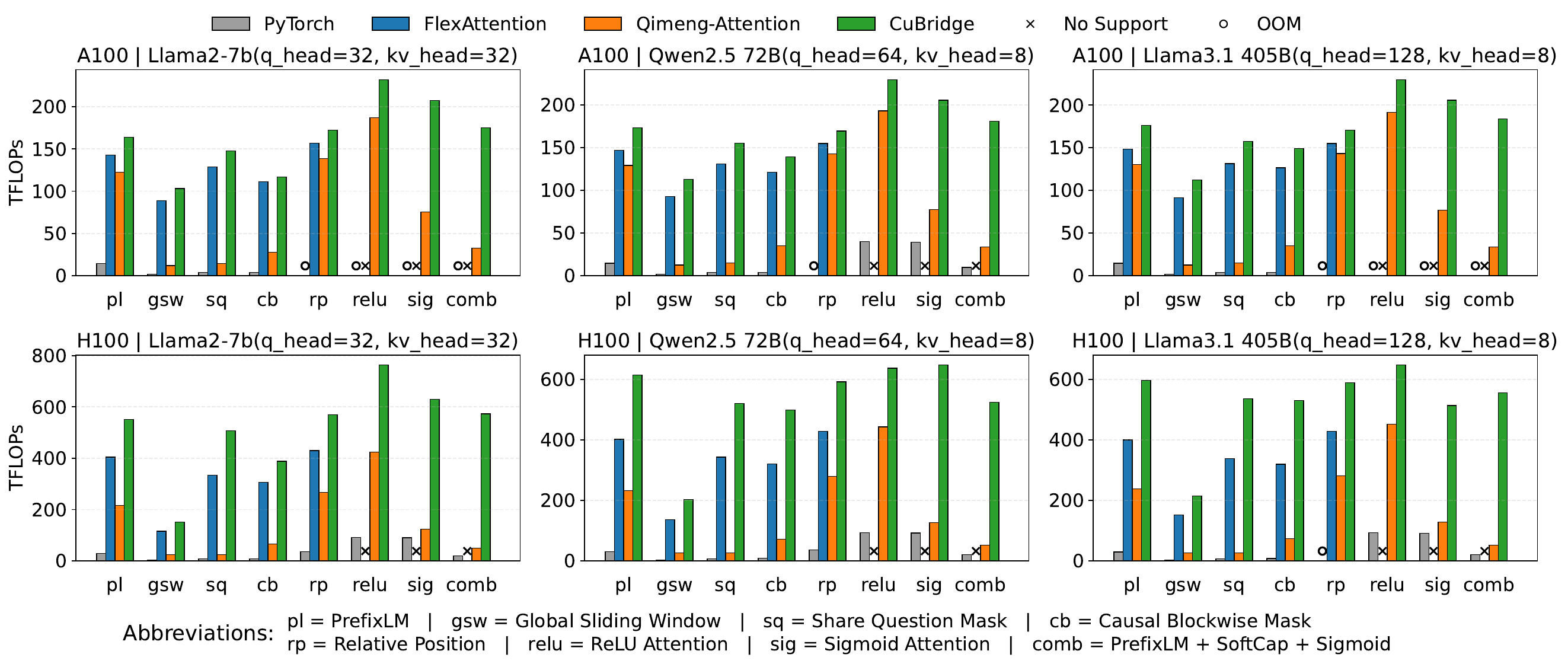}
    \vspace{-0.2cm}
    \caption{End-to-end performance comparison across attention variants and GPU platforms.}
    \vspace{-0.3cm}
    \label{fig:overall_performance}
\end{figure*}

\section{Experiment}
\label{sec:experiment}

\subsection{Experiment Setup}

We evaluate \proj{} along three aspects:
(1) end-to-end kernel correctness and performance across representative model configurations,
(2) robustness across diverse attention semantics, including previously unsupported variants,
and (3) the impact of key design choices via ablation studies.

\noindent\textbf{GPU platforms.}
All experiments are conducted on NVIDIA GPUs from two widely deployed architectures:
A100 (Ampere)~\cite{a100_whitepaper} and H100 (Hopper)~\cite{h100_whitepaper}.

\noindent\textbf{LLM Backends.}
We evaluate \proj{} using diverse SOTA LLMs, including the closed-source models of GPT-5~\cite{gpt5}, Claude-3.5-Sonnet~\cite{Claude}, as well as open-source models of DeepSeek-V3~\cite{deepseekV3}, \fixme{Qwen-3-235B~\cite{Qwen-3}, Qwen-3-32B~\cite{Qwen-3}}.

\noindent\textbf{Attention Variants.}
We focus our evaluation on attention mechanisms that are currently not supported in the standard FlashAttention~\cite{FlashAttention,FlashAttention-2,FlashAttention-3} library. 
Our test suite encompasses a diverse set of variants, including PrefixLM~\cite{Exploring_the_Limits}, Global Sliding Window~\cite{Longformer}, Share Question Mask~\cite{Leveraging_Passage_Retrieval}, Causal Blockwise Mask~\cite{Ring_Attention}, Relative Position Embeddings~\cite{Relative_Position}, ReLU Attention~\cite{FlashAttention}, and Sigmoid Attention~\cite{Sigmoid_Self-Attention}.
To assess robustness under unseen compositions, we additionally evaluate a composite variant that combines PrefixLM, Softcap, and Sigmoid attention.
All hyper-parameters follow the configurations reported in the original works.

\noindent\textbf{Comparison Baselines.}
\proj{} adopts FlashAttention v2.8.0 ~\cite{FlashAttention-2} as 
the source expert kernel reference.
\fixme{We compare CuBridge with three representative state-of-the-art baselines representing different paradigms currently explored in the literature}:
(1) PyTorch~\cite{pytorch} attention implementations built from standard operators;
(2) FlexAttention~\cite{FlexAttention}, a compiler-based framework for attention generation; 
and (3) {Qimeng-Attention}~\cite{QiMeng-Attention}, an LLM-based attention kernel generation approach.
\fixme{Additionally, we compare \proj{} against FlashInfer~\cite{flashinfer}, a high-performance expert-hand-tuned library.}

\noindent\textbf{Benchmark Configurations.}

We benchmark attention configurations derived from three representative real-world LLMs:
Llama2-7B (MHA, 32/32/128)~\cite{Llama2},
Qwen2.5-72B (GQA, 64/8/128)~\cite{Qwen2.5},
and Llama3.1-405B (GQA, 128/8/128)~\cite{Llamas3},
where the tuple denotes (query heads / key-value heads / head dimension).
Sequence length varies over $\{1\text{k}, 2\text{k}, 4\text{k}, 8\text{k}\}$, with batch size adjusted to maintain a constant total of 16k tokens per batch~\cite{FlashAttention}.
For all LLM-driven approaches, we set the generation temperature to zero~\cite{Kernelbench} and adopt a best-of-$k$ strategy with $k=10$~\cite{best-of-n}, reporting the best achieved performance.

\subsection{Overall Performance}

Figure~\ref{fig:overall_performance} compares the performance of \proj{} against all baselines on NVIDIA A100 and H100 GPUs.
We report results at sequence length $N=8\text{k}$ across diverse attention variants and model configurations.
Due to space constraints, the full results are provided in Appendix~\ref{sec:appendix_perf}.
Both \proj{} and the Qimeng-Attention baseline use GPT-5 as the underlying LLM backend.

Across all evaluated tasks, \proj{} supports all tested attention variants, achieves 100\% correctness, and consistently outperforms all state-of-the-art baselines.
On A100, \proj{} achieves average speedups of $12.69\times$, $1.18\times$ and $2.54\times$ over PyTorch, FlexAttention, and Qimeng-Attention, respectively;
on H100, these speedups further increase to $19.82\times$, $1.62\times$, $4.35\times$.
\fixme{Beyond these generative baselines, we also benchmark \proj{} against the hand-optimized FlashInfer library. 
\proj{} achieves comparable performance to FlashInfer on its natively supported variants (average 1.07$\times$) and substantial speedups (average 3.49$\times$) on variants beyond FlashInfer's native support. Detailed results and per-variant comparisons are provided in Appendix~\ref{sec:flashinfer}.}

Compared to PyTorch, \proj{} achieves significantly higher performance.
PyTorch realizes attention variants by composing multiple fine-grained operators into separate CUDA kernels, leading to frequent kernel launches, redundant global memory accesses, and occasional out-of-memory failures.
By generating a single fused kernel that integrates the full attention computation, \proj{} avoids these overheads and enables efficient data reuse.

Compared to FlexAttention, \proj{} demonstrates superior performance while supporting a broader range of attention variants.
As shown in Figure~\ref{fig:overall_performance}, \proj{} outperforms FlexAttention across all evaluated model configurations, with speedups of $1.05\times$--$1.22\times$ on A100 and $1.26\times$--$1.66\times$ on H100.
Notably, the performance gap widens on H100, indicating that \proj{} better adapts to newer GPU architectures.
This advantage stems from preserving expert-optimized, hardware-specific execution orchestration.
Beyond performance, \proj{} supports a broader set of attention variants that are not covered by FlexAttention, including ReLU Attention, Sigmoid Attention, and complex composite patterns, while maintaining high performance.

Compared to Qimeng-Attention, \proj{} consistently achieves higher performance across attention variants of varying complexity.
On A100, \proj{} attains speedups ranging from $1.18\times$ to $10.56\times$, while on H100 the speedups further increase to $1.43\times$--$20.3\times$.
As shown in Figure~\ref{fig:overall_performance}, on simpler variants such as relative position attention, \proj{} achieves performance comparable to Qimeng-Attention on A100 (average $1.24\times$ speedup).
In contrast, for variants with complex masking or irregular execution patterns (e.g., \texttt{share\_question\_mask} and \texttt{comb}), \proj{} delivers substantially larger gains, reaching $5.37\times$ on A100. 
For the same \texttt{comb} variants, speedups further increase to $11.47\times$ on H100, indicating that \proj{} better exploits the execution capabilities of newer GPU architectures.
By examining the generated kernels, we observe that while Qimeng-Attention introduces limited performance optimization, it does not construct explicit warp-specialized and tensor/cuda-core overlap execution orchestration.
These results suggest that relying solely on the LLM's internal knowledge makes it challenging to perform efficient execution orchestration for complex tasks, leading to the generated kernels only functional but inefficient.

\begin{figure}[t]
    \centering
    \includegraphics[width=\columnwidth]{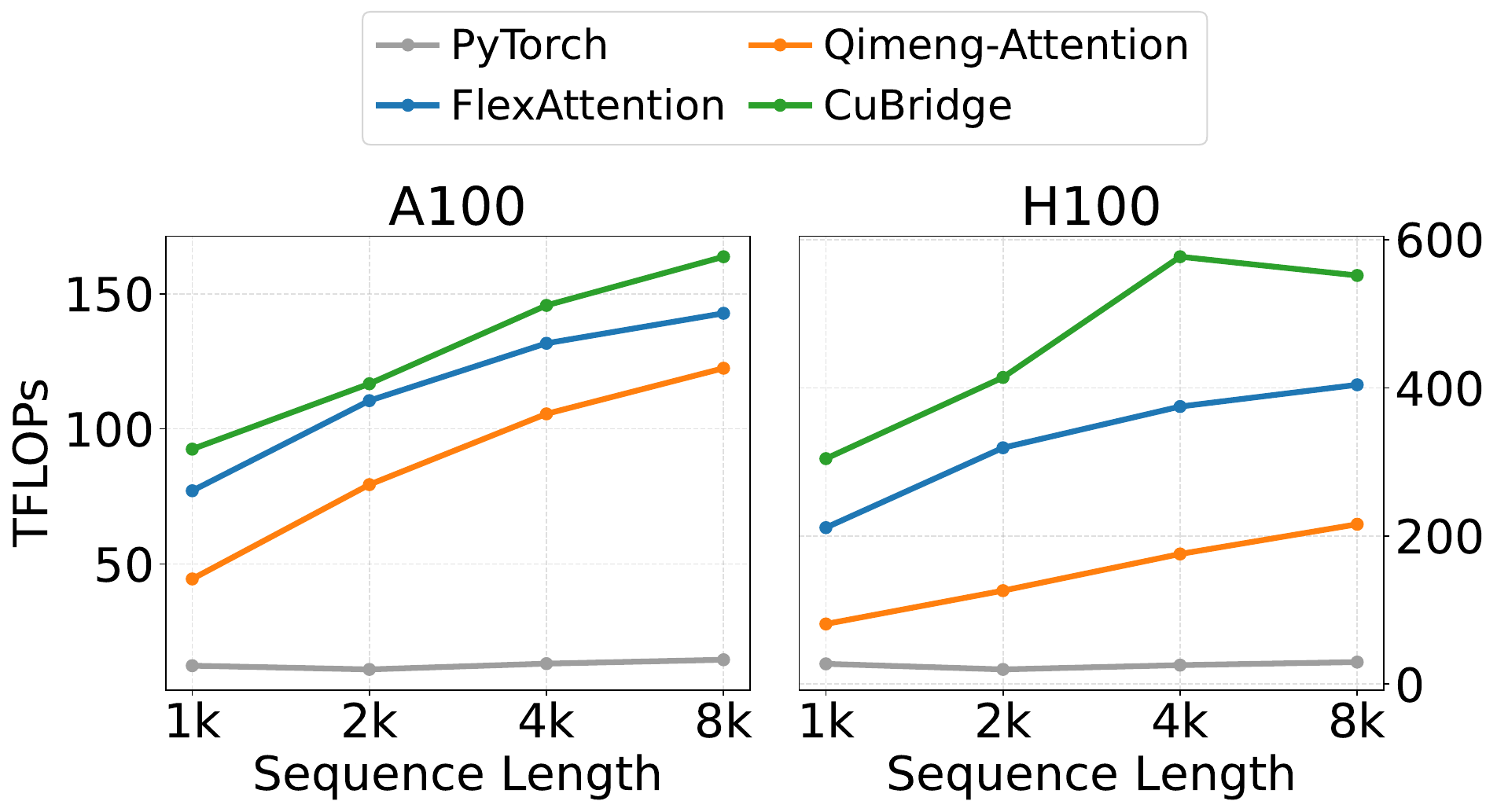}
    \vspace{-0.7cm}

    \caption{
    \fixme{Performance comparison for the PrefixLM variant (Llama2-7B config) across different sequence lengths on A100 (left) and H100 (right) GPUs.}}
    \label{fig:seqlen_vary_eval}
    \vspace{-0.5cm}
    
\end{figure}

Figure~\ref{fig:seqlen_vary_eval} reports performance across varying sequence lengths for the PrefixLM variant under the Llama2-7B configuration.
Across all evaluated sequence lengths, \proj{} consistently achieves higher throughput than SOTA baselines, indicating robust scalability with respect to sequence length.

\subsection{Ablation Study on System Components}

To validate the effectiveness of our system components, we conduct an ablation study on the task of rewriting expert-written FlashAttention kernels into target attention variants, guided by a user-provided PyTorch reference.
Experiments are conducted on H100 across 96 test cases (8 variants $\times$ 12 sequence lengths).
We compare three methods: 
(1) Vanilla GPT-5, which performs single-shot code rewriting of the source kernel based solely on the PyTorch reference;
(2) GPT-5 + ReAct, which applies iterative reasoning and code editing without \irname{};
and (3) \proj{}.
We report Pass@$k$ (whether at least one of $k$ samples passes correctness checking) and the geometric mean speedup of successful cases, normalized to Vanilla GPT-5.

Table~\ref{tab:ablation} presents the results.
Vanilla GPT-5 exhibits low success rates, which saturates quickly even with increased sampling (Pass@5 reaches only 0.38).
Adding ReAct reasoning improves correctness (Pass@1: 0.21 $\rightarrow$ 0.41), but the gain plateaus as $k$ increases and yields only limited speedup (1.23$\times$).
In contrast, \proj{} achieves perfect correctness (Pass@5 = 1.00), and substantially higher performance, with a 4.19$\times$ speedup.

This confirms that the structured IR provides essential support for reliable code modification. 
By lifting the source CUDA to the \irname{} level, \proj{} makes execution orchestration explicit and enables structured transformation and verification before lowering, leading to more stable correctness and higher performance.

\begin{table}[h]
\centering

\resizebox{0.48\textwidth}{!}{%
\begin{tabular}{lcccc}
\toprule
\textbf{Method} & \textbf{Pass@1} & \textbf{Pass@3} & \textbf{Pass@5} & \textbf{Norm. Speedup} \\
\midrule
GPT-5 & 0.21 & 0.33 & 0.38 & 1.00$\times$ \\
GPT-5 + ReAct & 0.41 & 0.54 & 0.58 & 1.23$\times$ \\
\textbf{GPT-5 + CuBridge} & \textbf{0.70} & \textbf{0.85} & \textbf{1.00} & \textbf{4.19$\times$} \\
\bottomrule
\end{tabular}%
}

\caption{Comparison of correctness and performance for different llm-based kernel modification methods across 96 test cases on H100. Performance is reported as the geometric mean of speedup normalized to the Vanilla GPT-5 baseline.}
\label{tab:ablation}

\vspace{-0.2cm}

\end{table}

\subsection{Impact of Different LLM Backends}

To evaluate the generality of \proj{}, we benchmark it using five representative LLM backends:
GPT-5, Claude, DeepSeek-V3, Qwen-3-235B and Qwen-3-32B. 
Table~\ref{tab:llm_impact} reports the achieved throughput (TFLOPS) on the H100 GPU for the PrefixLM task under the Llama2-7B attention configuration.

\fixme{The results demonstrate that \proj{} achieves stable, expert-level performance once the backend reaches a baseline CUDA reasoning capability. Specifically, substituting GPT-5 with Claude, DeepSeek-V3, or Qwen-3-235B results in less than 5\% performance variation across all sequence lengths. This indicates that our throughput gains primarily stem from the structured lift–transfer–lower workflow and \irname alignment, rather than reliance on a specific model.}

\fixme{However, we observe a distinct capability threshold for code reconstruction. While the larger models succeed, Qwen-3-32B fails to produce valid kernels. This suggests that reliable low-level CUDA generation requires a baseline level of reasoning ability. While different frameworks may vary in robustness, insufficient model capacity can become a limiting factor for complex kernel generation tasks.}

\begin{table}[h]
\centering

\resizebox{0.48\textwidth}{!}{%
\begin{tabular}{lrrrr}
\toprule
\textbf{LLM Backends} & \textbf{Seq=1k} & \textbf{Seq=2k} & \textbf{Seq=4k} & \textbf{Seq=8k} \\
\midrule
GPT-5 & \textbf{304.35} & 426.82 & \textbf{577.03} & 551.73 \\
Claude & 292.87 & \textbf{428.64} & {562.91} & \textbf{569.02} \\
DeepSeek-V3 & 294.12 & 424.05 & 557.03 & 549.73 \\
Qwen-3-235B	& 295.04 & 421.63 & 558.74 & 542.61 \\
Qwen-3-32B	& N/A & N/A & N/A & N/A \\
\bottomrule
\end{tabular}%
}

\caption{Performance comparison (TFLOPS) across different LLM backends on H100 (PrefixLM, Llama2-7b attention config).}
\label{tab:llm_impact}

\vspace{-0.2cm}

\end{table}

%% file: latex/conclusion.tex
\section{Conclusion}

We present \proj{}, an LLM-based framework for efficiently adapting expert-written attention kernels to new semantic variants.
\proj{} is built on \irname{}, an executable intermediate representation that makes execution orchestration explicit while abstracting low-level CUDA details, and uses a structured lift--transfer--lower workflow for reliable kernel adaptation.
Across diverse real-world attention variants on A100 and H100 GPUs, \proj{} consistently produces correct kernels and significantly outperforms general frameworks, template-based compilers, and prior LLM-based approaches.

%% file: latex/limitations.tex
\section*{Limitations}

Our work has two main limitations. 
First, \proj{} relies on the availability of high-quality expert kernels as source references.
While such optimized implementations are widely available in the NVIDIA platform, they are often lacking on non-mainstream customized hardware (e.g., FPGAs). 
This may limit the effectiveness of our framework in those environments.
Second, our current evaluation focuses primarily on attention variants, as they highlight the complexity of execution orchestration. 
We have not yet extended our validation to other high-performance computing tasks, such as scientific computing, which remains a promising direction for future work.

%% file: latex/acknowledgements.tex
\section*{Acknowledgements}

This work was supported by the Fundamental and Interdisciplinary Disciplines Breakthrough Plan of the Ministry of Education of China (No. JYB2025XDXM113), the National Natural Science Foundation of China (Nos. 62532006, 62572300), the Shanghai Qi Zhi Institute Innovation Program (SQZ202316), the Ministry of Education, Singapore (Nos. MOE-T2EP20124-0017, MOET32020-0004), and the 
CyberSG R\&D Cyber Research Programme Office (A-8002767-00-00).

Any opinions, findings, and conclusions in this paper are those of the authors only and do not necessarily reflect the views of our sponsors.

%% file: latex/appendix.tex
\section{Definitions of Attention Variants}
\label{app:attention_variants}

In this section, we provide the mathematical definitions and configurations for the attention variants used in our evaluation. 
To encompass various modifications to the scoring logic and normalization, we formulate the general attention computation as:
\begin{equation}
    O = \operatorname{\textbf{Norm}}\left( \operatorname{\textbf{Mask}}\left( \operatorname{\textbf{Score}}\left(\frac{QK^T}{\sqrt{d}}\right) \right) \right) V
\end{equation}
where $Q, K, V \in \mathbb{R}^{L \times d}$. The components are defined as follows:

{
\setlength{\parskip}{0pt}
\setlength{\itemsep}{0pt}
\begin{itemize}
    \setlength\itemsep{0em}
    \item $\operatorname{\textbf{Score}}(\cdot)$: Adjustments on raw attention scores (e.g., relative bias, softcapping).
    \item $\operatorname{\textbf{Mask}}(\cdot)$: Visibility constraints (e.g., causal, sliding window).
    \item $\operatorname{\textbf{Norm}}(\cdot)$: Normalization functions (e.g., Softmax, ReLU).
\end{itemize}
}
\noindent Unless otherwise specified, for the mask $M$, $M_{ij} = 0$ indicates visible, and $M_{ij} = -\infty$ indicates masked.

\subsection{Score Modifications}
Adjustments applied before masking. In the formulas below, $S_{ij} = q_i k_j^T / \sqrt{d}$ represents the raw attention score.

\paragraph{Relative Position Embeddings~\cite{Relative_Position}} 
Injects positional information by directly adding the relative index difference between the query ($i$) and key ($j$) to the score.
\[
\operatorname{\textbf{Score}}(S_{ij}) = S_{ij} + (i - j)
\]

\paragraph{Softcap Attention~\cite{gemma2}} 
Applies a `tanh` cap with a scaling factor $C$ to stabilize logits.
\[
\operatorname{\textbf{Score}}(S_{ij}) = C \cdot \tanh\left(\frac{S_{ij}}{C}\right)
\]

\subsection{Mask Variants}
These variants define the visibility pattern $M_{ij}$ between query $i$ and key $j$.

\paragraph{PrefixLM~\cite{Exploring_the_Limits}} 
Combines bi-directional attention for the prefix (length $L_p$) and causal attention for generated tokens.
\[
M_{ij} = \begin{cases} 
0 & \text{if } j \le L_p \\
  & \text{\scriptsize (Prefix visibility)} \\
0 & \text{if } L_p < j \le i \\
  & \text{\scriptsize (Causal generation)} \\
-\infty & \text{otherwise}
\end{cases}
\]

\paragraph{Global Sliding Window~\cite{Longformer}} 
Tokens attend to neighbors within a window $w$ and specific global tokens ($\mathcal{G}$).
\[
M_{ij} = \begin{cases} 
0 & \text{if } |i - j| \le w \\
0 & \text{if } i \in \mathcal{G} \text{ or } j \in \mathcal{G} \\
-\infty & \text{otherwise}
\end{cases}
\]

\paragraph{Share Question Mask~\cite{Leveraging_Passage_Retrieval}}
Commonly used in Reward Models and DPO, this mask allows multiple candidate responses (sets $A_k$) to attend to a single shared prompt ($Q$) to eliminate redundant computation.
\[
M_{ij} = \begin{cases}
0 & \text{if } i,j \in Q \text{ and } j \le i \\
  & \text{\scriptsize (Prompt self-attn)} \\
0 & \text{if } i \in A_k, j \in Q \cup A_k \text{ and } j \le i \\
  & \text{\scriptsize (Response $\to$ Prompt + Self)} \\
-\infty & \text{otherwise}
\end{cases}
\]

\paragraph{Causal Blockwise Mask~\cite{Ring_Attention}}
Isolates demonstration examples based on document IDs ($D$) while granting the test query (starting at $L_q$) visibility over the full context.
\[
M_{ij} = \begin{cases}
0 & \text{if } j \le i \text{ and } D_i = D_j \\
  & \text{\scriptsize (Demonstration isolation)} \\
0 & \text{if } j \le i \text{ and } i \ge L_q \\
  & \text{\scriptsize (Test query visibility)} \\
-\infty & \text{otherwise}
\end{cases}
\]

\subsection{Normalization Variants}
Replacements for the standard Softmax function. In these variants, normalization logic often uses sequence length $L$ for scaling.

\paragraph{ReLU Attention~\cite{FlashAttention}} 
Applies ReLU activation and scales by sequence length.
\[
\operatorname{\textbf{Norm}}(S) = \frac{1}{L} \operatorname{ReLU}(S)
\]

\paragraph{Sigmoid Attention~\cite{Sigmoid_Self-Attention}} 
Applies Sigmoid activation and scales by sequence length.
\[
\operatorname{\textbf{Norm}}(S) = \frac{1}{L} \operatorname{Sigmoid}(S)
\]

\begin{table*}[t]  
\centering

\resizebox{\textwidth}{!}{%
\begin{tabular}{llrrrrrrrrrrrr}
\toprule
 & & \multicolumn{4}{c}{\textbf{Llama2-7b}} & \multicolumn{4}{c}{\textbf{Qwen2.5 72B}} & \multicolumn{4}{c}{\textbf{Llama3.1 405B}} \\
 & & \multicolumn{4}{c}{\small ($q=32, k=32$)} & \multicolumn{4}{c}{\small ($q=64, k=8$)} & \multicolumn{4}{c}{\small ($q=128, k=8$)} \\
\cmidrule(lr){3-6} \cmidrule(lr){7-10} \cmidrule(lr){11-14}
\textbf{Task Variant} & \textbf{Method} & \textbf{1k} & \textbf{2k} & \textbf{4k} & \textbf{8k} & \textbf{1k} & \textbf{2k} & \textbf{4k} & \textbf{8k} & \textbf{1k} & \textbf{2k} & \textbf{4k} & \textbf{8k} \\
\midrule

 & Torch & 12.36 & 10.99 & 13.12 & 14.56 & 14.58 & 10.88 & 13.08 & 14.51 & 14.64 & 10.93 & 13.03 & 14.53 \\
 & FlexAttention & \underline{77.09} & \underline{110.44} & \underline{131.67} & \underline{142.77} & \underline{99.82} & \underline{121.96} & \underline{137.52} & \underline{146.68} & \underline{107.89} & \underline{128.63} & \underline{140.49} & \underline{148.55} \\
 & Qimeng Attn & 44.46 & 79.37 & 105.55 & 122.44 & 72.37 & 99.92 & 120.37 & 129.07 & 76.01 & 101.05 & 118.83 & 130.03 \\
\multirow{-4}{*}{\textbf{PrefixLM}} & \gc\textbf{CuBridge} & \gc\textbf{92.50} & \gc\textbf{116.69} & \gc\textbf{145.69} & \gc\textbf{163.70} & \gc\textbf{103.61} & \gc\textbf{138.70} & \gc\textbf{159.97} & \gc\textbf{172.98} & \gc\textbf{122.22} & \gc\textbf{149.36} & \gc\textbf{166.34} & \gc\textbf{176.39} \\
\midrule

 & Torch & 10.61 & 5.04 & 3.15 & 1.72 & 12.74 & 5.14 & 3.11 & 1.72 & 12.82 & 5.16 & 3.14 & 1.71 \\
 & FlexAttention & \underline{56.96} & \textbf{85.24} & \underline{87.67} & \underline{88.87} & \underline{92.52} & \underline{92.09} & \underline{92.32} & \underline{92.78} & \underline{96.40} & \underline{94.23} & \underline{92.47} & \underline{91.40} \\
 & Qimeng Attn & 29.82 & 28.11 & 19.88 & 11.86 & 47.55 & 34.92 & 22.14 & 12.43 & 49.95 & 35.44 & 21.81 & 12.46 \\
\multirow{-4}{*}{\shortstack[l]{\textbf{Global Sliding}\\\textbf{Window}}} & \gc\textbf{CuBridge} & \gc\textbf{67.36} & \gc\underline{80.77} & \gc\textbf{98.56} & \gc\textbf{103.05} & \gc\textbf{94.80} & \gc\textbf{102.50} & \gc\textbf{110.08} & \gc\textbf{112.75} & \gc\textbf{110.95} & \gc\textbf{111.42} & \gc\textbf{113.23} & \gc\textbf{112.23} \\
\midrule

 & Torch & 5.37 & 3.72 & 3.52 & 3.84 & 5.67 & 3.69 & 3.80 & 3.82 & 6.19 & 3.71 & 3.81 & 3.83 \\
 & FlexAttention & \underline{29.86} & \underline{84.32} & \underline{108.10} & \underline{128.78} & \underline{64.44} & \underline{92.66} & \underline{112.75} & \underline{130.71} & \underline{76.01} & \underline{95.16} & \underline{116.07} & \underline{131.24} \\
 & Qimeng Attn & 12.27 & 14.36 & 14.54 & 14.43 & 15.77 & 16.61 & 15.60 & 14.88 & 18.54 & 16.90 & 15.65 & 14.88 \\
\multirow{-4}{*}{\shortstack[l]{\textbf{Share}\\\textbf{Question Mask}}} & \gc\textbf{CuBridge} & \gc\textbf{39.81} & \gc\textbf{88.62} & \gc\textbf{122.31} & \gc\textbf{147.61} & \gc\textbf{73.96} & \gc\textbf{107.88} & \gc\textbf{135.07} & \gc\textbf{154.98} & \gc\textbf{90.18} & \gc\textbf{116.09} & \gc\textbf{138.86} & \gc\textbf{157.26} \\
\midrule

 & Torch & 4.85 & 3.67 & 3.89 & 4.00 & 6.31 & 3.72 & 3.86 & 3.97 & 6.33 & 3.75 & 3.89 & 3.98 \\
 & FlexAttention & \underline{31.77} & \textbf{62.96} & \textbf{87.36} & \underline{110.96} & \textbf{59.41} & \underline{79.85} & \underline{100.77} & \underline{120.91} & \underline{70.30} & \underline{87.45} & \underline{108.47} & \underline{126.37} \\
 & Qimeng Attn & 16.91 & 25.43 & 30.93 & 27.49 & 26.73 & 33.81 & 34.98 & 34.75 & 32.58 & 34.43 & 34.75 & 34.90 \\
\multirow{-4}{*}{\shortstack[l]{\textbf{Causal}\\\textbf{Blockwise Mask}}} & \gc\textbf{CuBridge} & \gc\textbf{35.12} & \gc\underline{54.27} & \gc\underline{86.21} & \gc\textbf{116.82} & \gc\underline{57.67} & \gc\textbf{85.10} & \gc\textbf{114.39} & \gc\textbf{139.24} & \gc\textbf{76.49} & \gc\textbf{102.34} & \gc\textbf{127.40} & \gc\textbf{148.91} \\
\midrule

 & Torch & 14.70 & 13.72 & 13.82 & OOM & 16.27 & 13.70 & 15.90 & OOM & 16.34 & 13.61 & 15.82 & OOM \\
 & FlexAttention & \textbf{114.85} & \underline{144.68} & \underline{149.75} & \underline{157.08} & \underline{131.96} & \underline{142.19} & \underline{152.02} & \underline{154.73} & \underline{134.26} & \underline{147.91} & \underline{152.28} & \underline{154.91} \\
 & Qimeng Attn & 71.05 & 112.10 & 130.56 & 138.79 & 105.91 & 126.59 & 139.74 & 142.81 & 109.04 & 127.91 & 135.62 & 143.22 \\
\multirow{-4}{*}{\textbf{Relative Pos.}} & \gc\textbf{CuBridge} & \gc\underline{105.80} & \gc\textbf{151.28} & \gc\textbf{162.47} & \gc\textbf{172.53} & \gc\textbf{137.92} & \gc\textbf{154.96} & \gc\textbf{166.21} & \gc\textbf{169.39} & \gc\textbf{146.46} & \gc\textbf{161.65} & \gc\textbf{167.97} & \gc\textbf{170.50} \\
\midrule

 & Torch & 31.16 & 36.59 & 33.63 & OOM & 32.35 & 36.44 & 38.28 & 39.98 & 33.79 & 37.38 & 39.73 & OOM \\
 & FlexAttention & \multicolumn{12}{c}{no support} \\
 & Qimeng Attn & \underline{88.56} & \underline{142.67} & \underline{140.83} & \underline{186.76} & \underline{134.64} & \underline{168.17} & \underline{186.99} & \underline{193.01} & \underline{137.73} & \underline{168.23} & \underline{182.67} & \underline{191.71} \\
\multirow{-4}{*}{\textbf{ReLU Attn}} & \gc\textbf{CuBridge} & \gc\textbf{150.88} & \gc\textbf{203.30} & \gc\textbf{219.46} & \gc\textbf{232.19} & \gc\textbf{181.02} & \gc\textbf{204.08} & \gc\textbf{225.00} & \gc\textbf{229.70} & \gc\textbf{192.73} & \gc\textbf{216.88} & \gc\textbf{225.96} & \gc\textbf{229.92} \\
\midrule

 & Torch & 30.48 & 35.99 & 31.95 & OOM & 30.94 & 35.80 & 37.46 & 39.14 & 33.37 & 36.70 & 38.91 & OOM \\
 & FlexAttention & \multicolumn{12}{c}{no support} \\
 & Qimeng Attn & \underline{45.05} & \underline{62.92} & \underline{66.33} & \underline{75.16} & \underline{54.36} & \underline{71.15} & \underline{74.58} & \underline{77.51} & \underline{64.54} & \underline{71.05} & \underline{74.98} & \underline{76.87} \\
\multirow{-4}{*}{\textbf{Sigmoid Attn}} & \gc\textbf{CuBridge} & \gc\textbf{134.56} & \gc\textbf{183.33} & \gc\textbf{196.13} & \gc\textbf{207.63} & \gc\textbf{165.47} & \gc\textbf{185.36} & \gc\textbf{201.14} & \gc\textbf{205.59} & \gc\textbf{173.81} & \gc\textbf{195.33} & \gc\textbf{201.92} & \gc\textbf{205.62} \\
\midrule

 & Torch & 8.33 & 9.41 & 9.46 & OOM & 9.11 & 9.38 & 9.45 & 9.62 & 9.18 & 9.50 & 9.63 & OOM \\
 & FlexAttention & \multicolumn{12}{c}{no support} \\
 & Qimeng Attn & \underline{17.78} & \underline{26.12} & \underline{29.39} & \underline{32.60} & \underline{25.46} & \underline{29.37} & \underline{31.83} & \underline{33.44} & \underline{26.90} & \underline{29.98} & \underline{32.13} & \underline{33.39} \\
\multirow{-4}{*}{\shortstack[l]{\textbf{Combo}\\\textbf{(PrefixLM+}\\\textbf{Softcap+}\\\textbf{Sigmoid)}}} & \gc\textbf{CuBridge} & \gc\textbf{80.03} & \gc\textbf{120.02} & \gc\textbf{155.65} & \gc\textbf{175.34} & \gc\textbf{106.20} & \gc\textbf{144.75} & \gc\textbf{167.16} & \gc\textbf{180.89} & \gc\textbf{124.73} & \gc\textbf{154.32} & \gc\textbf{172.60} & \gc\textbf{183.61} \\
\bottomrule
\end{tabular}%
}

\vspace{0.1cm} 
\caption{Performance benchmark (TFLOPS) of attention variants across varying sequence lengths and real-world model attention configurations on NVIDIA A100. (For each backbone the best result are marked in {bold}, and the second best result are {underlined}.)}
\label{tab:perf_a100}

\end{table*}

\begin{table*}[t] 
\centering

\resizebox{\textwidth}{!}{%
\begin{tabular}{llrrrrrrrrrrrr}
\toprule
 & & \multicolumn{4}{c}{\textbf{Llama2-7b}} & \multicolumn{4}{c}{\textbf{Qwen2.5 72B}} & \multicolumn{4}{c}{\textbf{Llama3.1 405B}} \\
 & & \multicolumn{4}{c}{\small ($q=32, k=32$)} & \multicolumn{4}{c}{\small ($q=64, k=8$)} & \multicolumn{4}{c}{\small ($q=128, k=8$)} \\
\cmidrule(lr){3-6} \cmidrule(lr){7-10} \cmidrule(lr){11-14}
\textbf{Task Variant} & \textbf{Method} & \textbf{1k} & \textbf{2k} & \textbf{4k} & \textbf{8k} & \textbf{1k} & \textbf{2k} & \textbf{4k} & \textbf{8k} & \textbf{1k} & \textbf{2k} & \textbf{4k} & \textbf{8k} \\
\midrule

 & Torch & 27.22 & 19.72 & 25.48 & 29.61 & 28.40 & 19.99 & 25.67 & 29.77 & 29.03 & 20.06 & 25.67 & 29.66 \\
 & FlexAttention & \underline{211.17} & \underline{319.03} & \underline{374.71} & \underline{404.09} & \underline{248.79} & \underline{333.70} & \underline{382.13} & \underline{402.38} & \underline{280.66} & \underline{335.76} & \underline{377.96} & \underline{400.05} \\
 & Qimeng Attn & 81.07 & 126.08 & 175.60 & 215.75 & 113.24 & 163.79 & 207.47 & 232.61 & 124.68 & 173.70 & 214.14 & 238.76 \\
\multirow{-4}{*}{\textbf{PrefixLM}} & \gc\textbf{CuBridge} & \gc\textbf{304.35} & \gc\textbf{414.05} & \gc\textbf{577.03} & \gc\textbf{551.73} & \gc\textbf{304.35} & \gc\textbf{486.19} & \gc\textbf{599.14} & \gc\textbf{615.26} & \gc\textbf{383.51} & \gc\textbf{548.49} & \gc\textbf{532.16} & \gc\textbf{597.61} \\
\midrule

 & Torch & 23.65 & 9.25 & 6.00 & 3.40 & 24.70 & 9.36 & 6.05 & 3.41 & 25.23 & 9.40 & 6.05 & 3.41 \\
 & FlexAttention & \underline{109.37} & \underline{111.62} & \underline{113.39} & \underline{115.70} & \underline{125.13} & \underline{136.57} & \underline{136.69} & \underline{136.88} & \underline{152.07} & \underline{153.13} & \underline{152.25} & \underline{152.01} \\
 & Qimeng Attn & 65.43 & 53.43 & 38.77 & 24.62 & 86.25 & 66.80 & 44.85 & 26.45 & 92.97 & 70.25 & 46.23 & 26.85 \\
\multirow{-4}{*}{\shortstack[l]{\textbf{Global Sliding}\\\textbf{Window}}} & \gc\textbf{CuBridge} & \gc\textbf{137.80} & \gc\textbf{176.88} & \gc\textbf{182.44} & \gc\textbf{151.90} & \gc\textbf{197.82} & \gc\textbf{214.90} & \gc\textbf{211.71} & \gc\textbf{202.74} & \gc\textbf{239.29} & \gc\textbf{249.16} & \gc\textbf{199.39} & \gc\textbf{214.46} \\
\midrule

 & Torch & 11.28 & 6.66 & 7.32 & 7.61 & 11.76 & 6.74 & 7.37 & 7.62 & 12.02 & 6.76 & 7.38 & 7.63 \\
 & FlexAttention & \underline{118.28} & \underline{188.58} & \underline{278.01} & \underline{333.53} & \underline{143.24} & \underline{224.59} & \underline{293.34} & \underline{343.28} & \underline{174.97} & \underline{272.52} & \underline{287.31} & \underline{337.92} \\
 & Qimeng Attn & 20.93 & 23.83 & 24.43 & 24.93 & 28.74 & 27.41 & 26.75 & 26.22 & 30.88 & 28.84 & 27.45 & 26.58 \\
\multirow{-4}{*}{\shortstack[l]{\textbf{Share}\\\textbf{Question Mask}}} & \gc\textbf{CuBridge} & \gc\textbf{205.51} & \gc\textbf{298.34} & \gc\textbf{398.49} & \gc\textbf{407.32} & \gc\textbf{176.67} & \gc\textbf{292.36} & \gc\textbf{437.08} & \gc\textbf{501.42} & \gc\textbf{286.09} & \gc\textbf{402.16} & \gc\textbf{422.20} & \gc\textbf{536.37} \\
\midrule

 & Torch & 11.31 & 6.77 & 7.56 & 8.10 & 12.10 & 6.84 & 7.62 & 8.14 & 12.46 & 6.88 & 7.63 & 8.10 \\
 & FlexAttention & \underline{115.00} & \textbf{179.56} & \underline{245.89} & \underline{306.93} & \underline{136.82} & \underline{202.30} & \underline{262.61} & \underline{321.27} & \underline{164.82} & \underline{213.50} & \underline{270.69} & \underline{319.44} \\
 & Qimeng Attn & 37.26 & 47.03 & 57.88 & 66.07 & 53.20 & 62.07 & 69.33 & 71.82 & 58.53 & 66.18 & 71.22 & 72.95 \\
\multirow{-4}{*}{\shortstack[l]{\textbf{Causal}\\\textbf{Blockwise Mask}}} & \gc\textbf{CuBridge} & \gc\textbf{157.27} & \gc\underline{177.58} & \gc\textbf{340.00} & \gc\textbf{388.39} & \gc\textbf{161.75} & \gc\textbf{277.11} & \gc\textbf{441.31} & \gc\textbf{498.62} & \gc\textbf{240.40} & \gc\textbf{366.02} & \gc\textbf{418.17} & \gc\textbf{530.28} \\
\midrule

 & Torch & 31.85 & 26.07 & 32.06 & 35.61 & 32.75 & 26.27 & 32.13 & 35.70 & 33.45 & 26.50 & 32.52 & -- \\
 & FlexAttention & \underline{301.57} & \underline{401.78} & \underline{420.14} & \underline{429.61} & \underline{328.40} & \underline{403.19} & \underline{416.66} & \underline{428.74} & \underline{334.10} & \underline{380.00} & \underline{407.89} & \underline{429.10} \\
 & Qimeng Attn & 143.41 & 199.26 & 242.67 & 266.63 & 184.73 & 231.65 & 264.98 & 279.89 & 193.11 & 238.75 & 265.39 & 281.31 \\
\multirow{-4}{*}{\textbf{Relative Pos.}} & \gc\textbf{CuBridge} & \gc\textbf{458.23} & \gc\textbf{527.63} & \gc\textbf{659.35} & \gc\textbf{568.84} & \gc\textbf{393.74} & \gc\textbf{519.15} & \gc\textbf{622.47} & \gc\textbf{592.46} & \gc\textbf{471.97} & \gc\textbf{595.68} & \gc\textbf{540.23} & \gc\textbf{589.83} \\
\midrule

 & Torch & 69.58 & 81.21 & 87.45 & 91.27 & 72.52 & 82.30 & 88.87 & 92.94 & 74.31 & 83.03 & 88.86 & 92.87 \\
 & FlexAttention & \multicolumn{12}{c}{no support} \\
 & Qimeng Attn & \underline{182.51} & \underline{280.51} & \underline{364.83} & \underline{424.78} & \underline{253.49} & \underline{346.74} & \underline{415.00} & \underline{443.19} & \underline{268.86} & \underline{357.82} & \underline{418.74} & \underline{451.79} \\
\multirow{-4}{*}{\textbf{ReLU Attn}} & \gc\textbf{CuBridge} & \gc\textbf{509.02} & \gc\textbf{605.51} & \gc\textbf{639.66} & \gc\textbf{764.12} & \gc\textbf{521.33} & \gc\textbf{632.60} & \gc\textbf{622.52} & \gc\textbf{637.52} & \gc\textbf{593.97} & \gc\textbf{615.01} & \gc\textbf{604.02} & \gc\textbf{648.54} \\
\midrule

 & Torch & 69.14 & 80.58 & 86.69 & 90.35 & 71.92 & 81.76 & 88.04 & 92.04 & 73.71 & 82.32 & 88.00 & 91.84 \\
 & FlexAttention & \multicolumn{12}{c}{no support} \\
 & Qimeng Attn & \underline{87.16} & \underline{105.95} & \underline{117.36} & \underline{123.61} & \underline{101.46} & \underline{115.04} & \underline{122.90} & \underline{127.11} & \underline{104.72} & \underline{117.21} & \underline{124.36} & \underline{128.04} \\
\multirow{-4}{*}{\textbf{Sigmoid Attn}} & \gc\textbf{CuBridge} & \gc\textbf{432.58} & \gc\textbf{584.86} & \gc\textbf{626.22} & \gc\textbf{630.25} & \gc\textbf{474.39} & \gc\textbf{626.13} & \gc\textbf{647.87} & \gc\textbf{648.71} & \gc\textbf{532.27} & \gc\textbf{635.80} & \gc\textbf{629.88} & \gc\textbf{514.04} \\
\midrule

 & Torch & 18.19 & 19.58 & 19.87 & 20.25 & 18.60 & 19.76 & 20.03 & 20.42 & 18.94 & 19.84 & 20.01 & 20.40 \\
 & FlexAttention & \multicolumn{12}{c}{no support} \\
 & Qimeng Attn & \underline{34.80} & \underline{40.97} & \underline{45.98} & \underline{49.91} & \underline{39.30} & \underline{44.81} & \underline{48.55} & \underline{51.39} & \underline{41.15} & \underline{45.94} & \underline{49.30} & \underline{51.90} \\
\multirow{-4}{*}{\shortstack[l]{\textbf{Combo}\\\textbf{(PrefixLM+}\\\textbf{Softcap+}\\\textbf{Sigmoid)}}} & \gc\textbf{CuBridge} & \gc\textbf{273.45} & \gc\textbf{411.90} & \gc\textbf{533.12} & \gc\textbf{572.94} & \gc\textbf{321.80} & \gc\textbf{490.90} & \gc\textbf{546.66} & \gc\textbf{524.12} & \gc\textbf{380.78} & \gc\textbf{532.47} & \gc\textbf{536.65} & \gc\textbf{555.81} \\
\bottomrule
\end{tabular}%
}

\vspace{0.1cm}
\caption{Performance benchmark (TFLOPS) of attention variants across varying sequence lengths and real-world model attention configurations on NVIDIA H100. (For each backbone the best result are marked in {bold}, and the second best result are {underlined}.)}
\label{tab:perf_h100}

\end{table*}

\section{Detailed Experimental Results}
\label{sec:appendix_perf}

This section provides the complete numerical data for the performance benchmarks discussed in Section~\ref{sec:experiment}. Table~\ref{tab:perf_a100} and Table~\ref{tab:perf_h100} detail the TFLOPS achieved by \proj{} and baselines across all evaluated attention variants, sequence lengths, and model attention configurations on NVIDIA A100 and H100 GPUs, respectively.

\begin{revision}
\section{Comparison With Expert Libraries}
\label{sec:flashinfer}
We compare \proj{} with FlashInfer(v0.6.0)~\cite{flashinfer} to evaluate its performance relative to expert-level manual optimization. 
FlashInfer is widely recognized as a performance benchmark due to its expert-optimized, hand-tuned kernels.

\subsection{Experiment Setup}
FlashInfer provides three distinct interfaces for implementing attention variants:
(1) Native APIs: Hand-optimized kernels for four built-in variants (\textit{Causal, Sliding Window, ALiBi, and Softcap}). 
(2) Block-Sparse (BSR) Wrapper: A block-sparse mask API (\texttt{BlockSparseAttentionWrapper}) that enables sparse mask but does not support score modification or non-Softmax normalization.
(3) JIT Templating: A Python JIT compilation API (\texttt{gen\_customize\_batch\_prefill\_module}) that allows users to inject custom C++ attention logic while keeping the execution schedule fixed.

All experiments were conducted on an NVIDIA H100 GPU using the Llama 3.1-405B attention configuration ($Q/KV$ heads: 128/8, Head dim: 128). We evaluate sequence lengths ranging from 1k to 8k, with the batch size dynamically adjusted to maintain a constant total of 16k tokens per batch.

For each evaluated variant, we benchmark all applicable FlashInfer interfaces and report the best FlashInfer performance (the maximum TFLOPS achieved among the available implementations) to ensure a rigorous comparison.

\subsection{Results and Discussion}
Table~\ref{tab:flashinfer_comp} presents the average throughput across sequence lengths from 1k to 8k. The results indicate that \proj{} matches the performance of native expert kernels ($0.99\times$--$1.22\times$) and significantly outperforms FlashInfer’s extension mechanisms on non-native variants ($2.73\times$--$5.59\times$).

For variants beyond FlashInfer’s native support, CuBridge achieves higher TFLOPS than the best FlashInfer configuration at every evaluated sequence length. 
The gap arises from how new semantics are realized. FlashInfer extensions attach additional semantics onto a fixed execution template without restructuring the schedule for the modified computation. 
Specifically, the BSR mechanism introduces extra global memory mask reads.
Moreover, while BSR enables block-level skipping, it does not support score transforms or non-softmax activations, and JIT allows semantic modification but fails to enable block-level sparse optimization, resulting in unnecessary computation. 
And neither BSR nor JIT currently supports H100-specific optimizations like warpgroup specialization, WGMMA, and TMA.
In contrast, CuBridge reconstructs kernels from a semantic specification via CuIR and re-derives the execution structure while preserving expert orchestration, enabling schedule-level adaptation for each variant.

\begin{table*}[t] 
\centering
\small 

\setlength{\tabcolsep}{8pt} 
\begin{tabular}{lccccc}
\toprule
\textbf{Attention Variant} & \textbf{FI (Native)} & \textbf{FI (BSR)} & \textbf{FI (JIT)} & \textbf{\proj{}} & \textbf{vs. Best FI} \\
\midrule
\gc Causal Mask & \textbf{551.62} & 110.95 & 176.54 & 546.18 & 0.99$\times$ \\
\gc Sliding Window & 276.12 & 78.92 & 38.27 & \textbf{282.60} & 1.02$\times$ \\
\gc Softcap & 383.90 & N/A & 91.20 & \textbf{466.85} & 1.22$\times$ \\
\gc ALiBi & 412.12 & N/A & 85.37 & \textbf{442.71} & 1.07$\times$ \\
\midrule
PrefixLM & N/A & 111.10 & 171.56 & \textbf{515.44} & 3.00$\times$ \\
Global Sliding Window & N/A & 82.61 & 68.28 & \textbf{225.25} & 2.73$\times$ \\
Share Question Mask & N/A & 99.28 & 11.89 & \textbf{411.71} & 4.15$\times$ \\
Causal Blockwise Mask & N/A & 125.00 & 35.67 & \textbf{388.72} & 3.11$\times$ \\
Relative Position & N/A & N/A & 163.61 & \textbf{549.43} & 3.36$\times$ \\
ReLU Attention & N/A & N/A & 197.06 & \textbf{615.39} & 3.12$\times$ \\
Sigmoid Attention & N/A & N/A & 158.25 & \textbf{578.00} & 3.65$\times$ \\
Combo & N/A & N/A & 89.76 & \textbf{501.43} & 5.59$\times$ \\
\bottomrule
\end{tabular}

\caption{Performance comparison (Avg TFLOPS) between \proj{} and FlashInfer (FI) on H100 GPU (Llama 3.1-405B configuration).}
\label{tab:flashinfer_comp}

\end{table*}

\section{Case Study: GEMM-based Operators}
\label{app:case_study_gemm}

To demonstrate the generality of \proj{} beyond attention mechanisms, we conducted an additional case study on GEMM-based operators. While attention was chosen as the primary focus due to its complexity, this study validates the broader applicability of our reference-driven paradigm.

Specifically, we use the open-source CUTLASS GEMM kernel as the source expert reference and apply \proj{} to generate a fused GEMM+ReduceSum variant. This transformation involves cross-thread accumulation and non-trivial epilogue modification. 

Table~\ref{tab:gemm_reduce} summarizes the results on H100 GPU. Across multiple common shapes, the generated GEMM+ReduceSum kernel consistently outperforms PyTorch, achieving 1.03$\times$--1.30$\times$ speedups. The performance improvement arises from kernel fusion: while PyTorch executes GEMM and Reduce-Sum as separate kernels, requiring intermediate results to be written to and read from global memory. In contrast, \proj{} directly modifies the expert reference kernel, enabling execution within a single fused kernel and eliminating intermediate global memory round-trips.
This case study demonstrates that \proj{}’s reference-driven transformation mechanism and \irname{} abstraction generalize to other core operator families.

\begin{table*}[t]
\centering
\small
\setlength{\tabcolsep}{8pt} 
\begin{tabular}{llccccc}
\toprule
\textbf{Implementation} & \textbf{Metric} & \textbf{(4k,4k,4k)} & \textbf{(1k,1k,1k)} & \textbf{(4k,4k,1k)} & \textbf{(2k,4k,1k)} & \textbf{(4k,2k,1k)} \\
\midrule
PyTorch & Latency (ms) & 0.2495 & 0.0319 & 0.0802 & 0.0505 & 0.0498 \\
        & TFLOPS       & 550.86 & 67.32  & 428.43 & 340.20 & 344.98 \\
\midrule
\textbf{\proj{} (Ours)} & \textbf{Latency (ms)} & \textbf{0.2425} & \textbf{0.0245} & \textbf{0.0580} & \textbf{0.0445} & \textbf{0.0384} \\
        & \textbf{TFLOPS}       & \textbf{566.76} & \textbf{87.65}  & \textbf{592.41} & \textbf{386.06} & \textbf{447.39} \\
\bottomrule
\end{tabular}

\caption{Performance comparison between PyTorch and \proj{} for the GEMM+ReduceSum operator on H100 GPU. Shape are denoted as $(M, N, K)$.}
\label{tab:gemm_reduce}

\end{table*}

\end{revision}

\section{Related Work}

\subsection{Current support for attention variants}

The surge in large-scale models has driven extensive GPU research, ranging from high-performance algorithms~\cite{VQ-LLM,ClusterFusion, zhou2021characterizing} and software stacks~\cite{FlashFuser, zhou2025sample} to innovative hardware utilization~\cite{guo2020balancing, hu2026m2xfp}. 
Within this landscape, the support for diverse attention variants has emerged as a particularly significant challenge, as it requires balancing algorithmic flexibility with the need for hardware-specific optimization.
Current systems for attention variants mainly fall into two categories. 

The first category is Expert Libraries.
Libraries like FlashAttention~\cite{FlashAttention} achieve state-of-the-art performance through heavy manual tuning.
However, they lack flexibility and only support a few variants (e.g., ALiBi, sliding window, causal mask, softcap).
Similarly, FlashInfer~\cite{flashinfer} applies these techniques to speed up LLM inference.
FlashMask~\cite{FlashMask} introduces a structural sparse mask representation, allowing it to support a wider range of mask variants.

The second category is {Compiler-based Approaches} (e.g., FlexAttention~\cite{FlexAttention}, AttentionEngine~\cite{AttentionEngine}). 
These methods improve programmability through template-based design.
Specifically, FlexAttention allows users to write simple PyTorch functions for score and mask computations. It then compiles these functions and injects them into fixed slots within a pre-written attention template.
While this offers some flexibility, it suffers from structural rigidity. Users cannot change logic outside these designated slots (e.g., they cannot replace the Softmax operator), which limits the types of algorithms it can support.
Moreover, the fixed execution pipeline prevents adaptation to hardware-specific opportunities like overlapping Tensor Core and CUDA Core workloads. As a result, these systems often still lag behind hand-tuned expert libraries in performance.

\subsection{LLM for CUDA Code Generation}
Recently, with the continuous improvement of LLMs, there is growing interest in using them to automatically generate GPU kernels.
However, SOTA models perform poorly on benchmarks like KernelBench~\cite{Kernelbench} and TritonBench~\cite{tritonbench}. 
This stems from the complexity of parallel programming models (e.g., managing hierarchical execution units and memory system) combined with the {extreme rarity of high-quality GPU code} in pre-training corpora (less than 0.1\% in The Stack~\cite{stack}).
{Existing research generally falls into two categories.}

\textbf{Training-based Methods.} 
These approaches try to improve the model's mastery of CUDA code through additional training. 
Methods like Kevin~\cite{Kevin} and CUDA-L1~\cite{CUDA-L1} apply reinforcement learning to this domain, using generated samples for training to help with the lack of data. 
Additionally, QiMeng-MuPa~\cite{QiMeng-MuPa} uses a mutual supervision framework where a code generator and a test case generator improve together. 
However, these methods require high training costs to modify the underlying model. {Our framework is orthogonal to training-based approaches and can be used in conjunction with them.}

\textbf{Workflow-based Methods.} 
{This category treats the LLM as a modular component within a system and focuses on the design of systemic workflows to complete complex tasks.~\cite{Astra, nvidia_blog}. }
Some use execution-guided feedback pipelines~\cite{nvidia_blog, CUDA-LLM}, while others employ multi-agent frameworks (such as Astra~\cite{Astra}, CUDA-LLM~\cite{CUDA-LLM}, and CudaForge~\cite{CudaForge}) that separate planning, generation, and debugging into distinct roles. 
Similar pipelines have also been explored for Triton kernel generation~\cite{KernelAgent, QiMengkernel}. 
However, these frameworks typically treat code as plain text, lacking insight into the complex execution orchestration required for performance. 
In contrast, \proj{} uses a structured IR to {make execution orchestration explicit while abstracting low-level CUDA details and use lift--transfer--lower workflow}. 
This enables the model to understand and modify the kernel's logical structure and data flow, ensuring that complex orchestration patterns are correctly handled.

\section{Lifter Agent System Prompt}
\label{app:system_prompt}

The complete system prompt of Lifter Module is presented below.

\begin{columnprompt}
    \textbf{\# Role}\\
    You are an expert CUDA System Engineer specializing in \textbf{Semantic Lifting}.
    Your goal is to transform low-level, expert-optimized CUDA kernels into \textbf{CuIR}, a Pythonic IR designed to capture \textbf{Execution Orchestration} while abstracting away implementation complexity.
    
    \vspace{4pt}
    \noindent\rule{\linewidth}{0.4pt} 
    \textbf{\# Task: The Semantic Lifting Process}\\
    You must process the CUDA kernel \textbf{step-by-step} to ensure semantic reliability:
    \begin{enumerate}[leftmargin=*, nosep]
        \item \textbf{Step 1: Low-level Annotation.} Begin by commenting all \code{ptx} usages and \code{CuTe} APIs. You must refer to the technical document to explain the exact hardware behavior of each instruction.
        \item \textbf{Step 2: Worker Identification.} Analyze the kernel's parallel hierarchy to identify distinct execution units (e.g., warp, warpgroup, threadblock) and their code snippet.
        \item \textbf{Step 3: CuIR construction.} Before writing any CuIR, reason about how these workers are synchronized. Identify the specific barriers and arrival counts that manage the execution pipeline. Based on the analysis above, map the logic into \irname{} primitives. You must ensure that the lifted representation strictly preserves the original asynchronous dataflow and synchronization logic.
    \end{enumerate}
    
    \noindent\rule{\linewidth}{0.4pt} 
    
    \textbf{\# CuIR Specification}
    
    \textbf{\#\# 1. Runtime System}\\
    The runtime defines the virtual execution environment for the IR.
    
    \begin{description}
        \item[\code{class KernelBase}] Represents the context of a {Single Thread Block}.
        \begin{itemize}
            \item \code{\_\_init\_\_(self, *args)}: Initialize primitives and allocate Shared Memory here.\\
            {Critical}: You MUST calculate \code{self.gridDim} (e.g., \code{M // BM}) based on input shapes to configure the simulator.
            \item \code{self.blockIdx}: A \code{Dim3} object mimicking CUDA's \code{blockIdx}.
            \item \code{bind(self, func, role\_name)}: Control Primitive. Maps a Python method to a virtual hardware worker.
            \item \code{commit(self)}: Control Primitive. Triggers the parallel execution of bound workers.
        \end{itemize}
    \end{description}

    \textbf{\#\# 2. CuIR Primitives}
    
    \textbf{\#\#\#  Memory \& Data Movement}
    \begin{itemize}
        \item \code{alloc(tile\_shape, hierarchy, dtype)}:\\
        Allocates a memory tile on the specified \code{hierarchy} (e.g., "shared", "register").\\
        \textit{Returns}: A Tensor-like object that supports direct indexing, slicing, and standard PyTorch math operations.
        
        \item \code{copy\_async(dst, src, barrier, instruction)}:\\
        Performs an \textbf{asynchronous copy} from \code{src} to \code{dst}.
        \begin{itemize}[leftmargin=1em]
            \item The \code{barrier} will be updated when transaction completion.
            \item \code{instruction}: (String) Annotation tag (e.g., "cp.async", "tma"). Purely for Semantic Lifting info; does not affect simulation logic.
        \end{itemize}
        
        \item \code{copy(dst, src, instruction)}:\\
        Performs a \textbf{synchronous copy} from \code{src} to \code{dst}.
        \begin{itemize}[leftmargin=1em]
            \item \code{instruction}: (String) Annotation tag (e.g., "ldmatrix", "stsw"). Purely for Semantic Lifting info.
        \end{itemize}
    \end{itemize}
    
    \vspace{2pt}
    \textbf{\#\#\#  Compute}
    \begin{itemize}
        \item \code{gemm\_async(A, B, instruction)}:\\
        Performs an \textbf{asynchronous matrix multiplication} (\code{A @ B}).
        \begin{itemize}[leftmargin=1em]
            \item Synchronization: Execution logic assumes results are not immediately available; completion MUST be managed via \code{wait\_group} primitives.
            \item \code{instruction}: (String) Annotation tag (e.g., "wgmma.mma\_async"). Purely for Semantic Lifting info.
        \end{itemize}
        
        \item \code{gemm(A, B, instruction)}:\\
        Performs a \textbf{synchronous matrix multiplication} (\code{A @ B}).
        \begin{itemize}[leftmargin=1em]
            \item \code{instruction}: (String) Annotation tag (e.g., "mma.sync"). Purely for Semantic Lifting info.
        \end{itemize}
        
        \item {CUDA Core Operations}:\\
        For standard ALU/SFU operations (Add, Mul, Exp, etc.), use native PyTorch operators directly on the Tensors (e.g., \code{C = A + B}, \code{torch.exp(x)}).
    \end{itemize}
    
    \vspace{2pt}
    \textbf{\#\#\#  Synchronization}
    
    Models the \code{mbarrier} hardware primitive. It tracks two distinct goals: \textbf{Thread Arrivals} (persistent) and \textbf{Transaction Bytes} (transient).
    \begin{itemize}
        \item \code{barrier.init(expected\_threads)}:\\
        Sets the persistent \code{base\_goal} (number of participating threads).
        \begin{itemize}[leftmargin=1em]
            \item Call this once in \code{\_\_init\_\_} or before the main loop.
        \end{itemize}
        
        \item \code{barrier.arrive\_expect\_tx(bytes\_count)}:\\
        \begin{enumerate}[leftmargin=1.5em]
            \item Signals thread arrival (increments \code{current\_threads}).
            \item Increases the transaction goal by \code{bytes\_count} (increments \code{tx\_goal}).
        \end{enumerate}
        Use this immediately before issuing a \code{copy\_async} (TMA) operation.
        
        \item \code{barrier.arrive()}:\\
        \begin{itemize}[leftmargin=1em]
             \item Signals thread arrival (increments \code{current\_threads}) without modifying transaction expectations.
        \end{itemize}
        
        \item \code{barrier.wait(parity)}:
        \begin{itemize}[leftmargin=1em]
            \item Blocks the thread until the barrier completes the phase associated with \code{parity}.
        \end{itemize}
    \end{itemize}
    
    \noindent\rule{\linewidth}{0.4pt}
    
    \textbf{\# Lifting Rule}
    
    \textbf{\#\#\# 1. Worker-Logic Disentanglement}
    \begin{itemize}
        \item Role Identification: Analyze control flow (e.g., \code{if warp\_group\_idx == 0}) to distinguish roles.
        \item Logic Merging: If multiple hardware units perform the same logical task, merge them into a single logic method.
    \end{itemize}
    
    \textbf{\#\#\# 2. Logical Tile Allocation (Memory)}
    \begin{itemize}
        \item Shared Memory: Allocate in \code{\_\_init\_\_} using \code{alloc(scope="shared")}.
        \item Register File: When allocating accumulators, derive the shape from the Logical Tile View (e.g., \code{[BM, BN]}) rather than the fragmented thread-level view.
    \end{itemize}
    
    \textbf{\#\#\# 3. Preservation of Execution Orchestration (Crucial)}
    \begin{itemize}
        \item Instruction Scheduling: You MUST preserve the exact relative order of Memory, Compute, and Barriers.
        \item Latency Hiding: If the CUDA source interleaves computation with pipeline stages, you must replicate this sequence 1:1. Do not reorder instructions.
    \end{itemize}
    
    \noindent\rule{\linewidth}{0.4pt}

     \textbf{\# Example}\\
     \textbf{...}
     
    \noindent\rule{\linewidth}{0.4pt} 
    \textbf{\# Output Format (Strict)}\\
    Your output MUST be a single executable Python code block.
    \begin{enumerate}[leftmargin=*, nosep]
        \item Define one Python class inheriting from \code{KernelBase}.
        \item Implement \code{\_\_init\_\_}, \code{run}, and logic methods.
        \item Do NOT include explanations or markdown text outside the code block.
    \end{enumerate}
\end{columnprompt}

\begin{revision}

\begin{table*}[b]
\centering
\small
\setlength{\tabcolsep}{10pt}
\renewcommand{\arraystretch}{1.2}

\begin{tabular}{lp{5cm}p{9cm}}
\toprule
\textbf{Rule} & \textbf{Transformation Category} & \textbf{Core Action} \\
\midrule
R1 & Iteration domain transformation & Modify loop bounds/iteration space while preserving tiling. \\
R2 & Execution structure preservation & Retain async pipeline and scheduling patterns. \\
R3 & Computation injection or modification & Insert or adapt intermediate computations. \\
R4 & Reduction or normalization replacement & Modify accumulation or normalization strategy. \\
R5 & Data dependency extension & Introduce additional inputs or memory staging. \\
R6 & Auxiliary structure adaptation & Adjust code logic to match control-flow changes. \\
R7 & Unaffected region preservation & Preserve IR regions not implicated by semantic delta. \\
\bottomrule
\end{tabular}

\caption{Taxonomy of IR Transformation Rules. These rules constrain the LLM to preserve the expert execution structure while adapting to new attention semantics.}
\label{tab:transfer_rules}
\vspace{-0.2cm}

\end{table*}

\section{IR Transfer Implementation Details}
\label{app:ir_transfer}

IR Transfer is implemented as an iterative transformation-and-validation loop. Given an Origin \irname{}, an Origin PyTorch specification, and a Target PyTorch specification, \proj{} generates a Target \irname{} under explicit transformation constraints, validates it against the Target PyTorch reference, and iteratively refines it if needed. The core logic is defined in the following procedure:

The generation stage is LLM-driven and explicitly rule-constrained. Similar to the lifting stage, IR Transfer uses a structured chain-of-thought prompt with ordered reasoning steps. The LLM is instructed to: (1) identify the semantic delta between the Origin and Target PyTorch specifications, (2) localize the affected region in the Origin \irname{}, and (3) apply predefined transformation rules to construct the Target \irname{}. 

The explicit rule taxonomy, as detailed in Table~\ref{tab:transfer_rules}, serves two purposes: it constrains modifications to prevent unintended disruption of expert scheduling structure, and it makes the transfer procedure compositional and reproducible rather than free-form code synthesis. After generation, verification is performed via the \irname{} runtime before CUDA reconstruction. Failed cases trigger structured diagnostics and bounded refinement.

\begin{tcolorbox}[
    colback=boxback, 
    colframe=boxframe, 
    arc=1mm, 
    title=Algorithm: IR Transfer Loop, 
    fonttitle=\bfseries\small, 
    boxsep=1pt,
    left=1pt,
    right=1pt,
    top=1pt,
    bottom=1pt,
    titlerule=0pt
]
\begin{lstlisting}[
    language=Python, 
    basicstyle=\footnotesize\ttfamily, % 从 \tiny 上调至 \footnotesize
    breaklines=true,
    columns=flexible % 增加此项可以让字符间距更自然
]
target_cuir = GenerateTargetCuIR(origin_cuir, origin_pytorch, target_pytorch)
is_valid, diagnostics = VerifyAgainstReference(target_cuir, reference=target_pytorch)

while not is_valid and iteration < max_iterations:
    target_cuir = RefineCuIR(target_cuir, diagnostics)
    is_valid, diagnostics = VerifyAgainstReference(target_cuir, reference=target_pytorch)
    iteration += 1
\end{lstlisting}
\end{tcolorbox}

\end{revision}